\documentclass[runningheads]{llncs}

\usepackage{eccv}

\usepackage{eccvabbrv}

\usepackage{graphicx}
\usepackage{booktabs}
\usepackage{algorithmic}
\usepackage[accsupp]{axessibility}  

\usepackage{hyperref}

\usepackage{orcidlink}

\usepackage{kotex}
\usepackage{enumitem}
\usepackage{amsmath}
\usepackage{multirow} 
\usepackage[ruled,vlined, linesnumbered]{algorithm2e}
\usepackage{pifont}
\makeatletter
\newcommand{\printfnsymbol}[1]{%
  \textsuperscript{\@fnsymbol{#1}}%
}
\makeatother

\begin{document}

\title{Mixed Non-linear Quantization \newline for Vision Transformers} 

\titlerunning{Mixed Non-linear Quantization \newline for Vision Transformers}



\author{Gihwan Kim\inst{1}\printfnsymbol{4}\orcidlink{0009-0006-8659-8518} \and
Jemin Lee\inst{2}\printfnsymbol{4}\orcidlink{0000-0002-9332-3508} \and
Sihyeong Park\inst{3}\orcidlink{0000-0001-8244-4817} \and  
Yongin Kwon\inst{2}\printfnsymbol{6}\orcidlink{0000-0003-2973-246X} \and
Hyungshin Kim\inst{1}\printfnsymbol{6}\orcidlink{0000-0001-9615-1644}}


\authorrunning{G.~Kim and J.~Lee \textit{et al.}}

\institute{Chungnam National University \\
\email{gihwan.kim98@o.cnu.ac.kr and hyungshin@cnu.ac.kr}\and
Electronics and Telecommunications Research Institute \\
\email{\{leejaymin,yongin.kwon\}@etri.re.kr} \and
Korea Electronics Technology Institute \\
\email{sihyeong@keti.re.kr}}


\maketitle

\def\thefootnote{\printfnsymbol{4}}\footnotetext{These authors contributed equally to this work and are listed alphabetically.}\def\thefootnote{\arabic{footnote}}
\def\thefootnote{\printfnsymbol{6}}\footnotetext{Corresponding authors}\def\thefootnote{\arabic{footnote}}

\begin{abstract}
The majority of quantization methods have been proposed to reduce the model size of Vision Transformers, yet most of them have overlooked the quantization of non-linear operations.
Only a few works have addressed quantization for non-linear operations, but they applied a single quantization method across all non-linear operations.
We believe that this can be further improved by employing a different quantization method for each non-linear operation. Therefore, to assign the most error-minimizing quantization method from the known methods to each non-linear layer, we propose a mixed non-linear quantization that considers layer-wise quantization sensitivity measured by SQNR difference metric.  
The results show that our method outperforms I-BERT, FQ-ViT, and I-ViT in both 8-bit and 6-bit settings for ViT, DeiT, and Swin models by an average of 0.6\%p and 19.6\%p, respectively. 
Our method outperforms I-BERT and I-ViT by 0.6\%p and 20.8\%p, respectively, when training time is limited.
We plan to release our code\footnote{\url{https://gitlab.com/ones-ai/mixed-non-linear-quantization}}.
\keywords{Quantization \and Vision Transformer \and Non-linear Quantization}
\end{abstract}

\section{Introduction}
\label{sec:intro}

Vision Transformers have replaced traditional Convolutional Neural Networks (CNNs) in various vision tasks due to their high accuracy. Alongside this, research on quantization to reduce model size for deployment on diverse devices, including resource-constrained devices, are actively ongoing. Quantization methods studied for ViTs are divided into quantization-aware training (QAT)~\cite{li2022qNeurips,li2022q,liu2023oscillation,huang2023variation,li2024bi,yuan2024vit} which involves learning, and post-training quantization (PTQ)~\cite{liu2021post,yuan2022ptq4vit,apq22,li2023repq,zhong2023s,zhongerq} which does not require learning. However, these studies have applied quantization only to the linear operations, neglecting many non-linear operations such as Softmax, GELU, and layer normalization (LayerNorm). 
Consequently, during inference, some or all operations utilize dequantized floating-point parameters, failing to take full advantage of efficient low-precision arithmetic units and thus leading to insufficient acceleration of the model.

To address these issues, studies on  quantizing the non-linear operations in ViT models have been recently proposed~\cite{pmlr-v139-kim21d, Lin2021FQViTPQ, dong2024packqvit, Lin2021FQViTPQ, zhong2023s, li2023repq, Li_2023_ICCV, wang2023sole}.
However, these methods use polynomial~\cite{pmlr-v139-kim21d, Lin2021FQViTPQ, dong2024packqvit}, logarithm~\cite{Lin2021FQViTPQ, zhong2023s, li2023repq}, and bit-shifting~\cite{Li_2023_ICCV} techniques to transform all non-linear operations into linear operations and perform the same quantization for all non-linear operations in ViTs.
However, the output activation distribution of non-linear operations varies widely depending on the position of the operators.
Consequently, uniformly applying a single non-linear quantization method to the entire non-linear operations is suboptimal in terms of minimizing quantization errors.

In this paper, we propose a mixed non-linear quantization approach that considers layer-wise quantization sensitivity within vision transformers. We utilize multiple quantization methods and we assign the best quantization method for each non-linear operation. Our method selects a quantization approach with the least quantization error according to the characteristics of the non-linear operation and the position of each non-linear layer. To measure the quantization error on a layer-wise basis, we devised a new metric called \textit{SQNR diff}. This metric represents the difference in SQNR values between input and output activations. This allows the assignment of non-linear quantization methods based on improvements in SQNR.

The considered non-linear quantization methods include three methods proposed in I-BERT~\cite{pmlr-v139-kim21d}, FQ-ViT~\cite{Lin2021FQViTPQ}, and I-ViT~\cite{Li_2023_ICCV}: bit-shifting~\cite{Li_2023_ICCV}, logarithm~\cite{Lin2021FQViTPQ}, and polynomial~\cite{pmlr-v139-kim21d}. Quantizing vision transformers with our mixed non-linear quantization approach reduces quantization error in non-linear layers on a layer-wise basis, allowing for more precise quantization than previous works. When performing layer-wise non-linear operation analysis, mixed non-linear quantization enhanced quantization sensitivity compared to existing quantization methods. Furthermore, when measuring top-1 accuracy for image classification on various vision transformer models, the accuracy was higher than that of existing integer-only vision transformer models.
The proposed method improved accuracy by an average of 0.6\%p and 19.6\%p in 8-bit and 6-bit environments for ViT, DeiT, and Swin models compared to the conventional studies I-BERT, FQ-ViT, I-ViT. Notably, even when quantization was performed with only 1-epoch QAT, it resulted in a mere -1.33\%p accuracy difference compared to full training, a significantly different result from the -22.27\%p average accuracy drop caused by I-ViT in full training comparisons.

The main contributions of this paper are as follows:
\begin{itemize}[noitemsep]
  \item We discover that applying a single quantization method to non-linear operations in ViT models does not minimize quantization errors effectively. This observation is based on layer-wise \textit{SQNR diff } analysis, which demonstrate that a single quantization approach often fails to address the diverse activation distribusion of different non-linear layers.
  \item We propose a mixed non-linear quantization method that combines three existing quantization methods—bit-shifting, logarithm, and polynomial quantization—to reduce quantization error in each non-linear operation.
  \item Our method achieves higher accuracy compared to previous works including I-BERT, FQ-ViT, and I-ViT. Furthermore, even with only one epoch of retraining, it attains higher accuracy than prior studies.

\end{itemize}

\section{Related Work}
\label{sec:Related Work}
\subsection{Vision Transformer Quantization}
\label{sec:Vision Transformer Quantization}
Recent studies have actively pursued the quantization of Vision Transformer (ViT) models to compress them~\cite{gholami2022survey,du2024model}.
Research on quantizing ViT models can be divided into the PTQ and the QAT methods.
In the PTQ, quantization have been proposed to address the increase in quantization error due to the wide distribution of activations~\cite{liu2021post,yuan2022ptq4vit,apq22,li2023repq,zhong2023s,zhongerq}.
QAT methods~\cite{li2022qNeurips,li2022q,liu2023oscillation,huang2023variation,li2024bi,yuan2024vit} enable quantization below 4-bit thanks to retraining, unlike PTQ.
However, these ViT works have applied quantization only to linear operations existing in ViTs, and many non-linear operations like Softmax have not been quantized.
Therefore, throughout the inference process, specific operations are carried out using floating-point parameters that have been converted to a continuous scale, so not fully using the capabilities of high-performance low-precision arithmetic units and resulting in inefficient acceleration of the model.

To address the aforementioned issues, recent studies have proposed integer-only vision transformer models that also quantize all non-linear operations, as listed in Table~\ref{table:non-linear approximation functions}~\cite{Li_2023_ICCV, Lin2021FQViTPQ, dong2024packqvit, wang2023sole, zhang2023practical}.
I-ViT~\cite{Li_2023_ICCV} performed quantization by approximating the operations of Softmax, LayerNorm, and GELU to integer operations using bit-shifting.
FQ-ViT~\cite{Lin2021FQViTPQ} used log2 quantization and \textit{i-exp}~\cite{pmlr-v139-kim21d} to quantize LayerNorm and Softmax.
PackQViT~\cite{dong2024packqvit} quantized non-linear operations using polynomial approximation.
However, existing non-linear quantization methods have only quantized some non-linear operations~\cite{Lin2021FQViTPQ} or work exclusively on custom-designed hardware, thus lacking versatility~\cite{wang2023sole}.
Furthermore, fundamentally, these methods are not optimal in minimizing quantization errors because they uniformly apply a non-linear quantization approach to all non-linear operations within the model, failing to quantize according to the distribution of activations~\cite{Li_2023_ICCV,Lin2021FQViTPQ,dong2024packqvit,pmlr-v139-kim21d,zhang2023practical}.

\subsection{Mixed Quantization}
\label{sec:Mixed Quantization}
Mixed-precision quantization is a quantization technique that assigns different bit-widths to each layer or block of a Neural Network considering the trade-off between accuracy and efficiency. By variably assigning bit-widths per layer, it can reduce accuracy degradation compared to fixed-precision quantization~\cite{gholami2022survey}. The method of HAQ~\cite{wang2019haq} uses Deep Reinforcement Learning to decide bit-widths. To determine bit-widths, HAWQ~\cite{dong2019hawq} and HAWQ-V2~\cite{dong2020hawq} respectively use the top eigenvalue and average hessian trace to measure layer-wise sensitivity to quantization. Furthermore, HAWQ-V3~\cite{yao2021hawq} has implemented models with mixed bit-widths for integer-only quantization. 
Recently, applying mixed bit-width quantization to vision transformers~\cite{liu2021post} optimized the bit-width allocation per layer by considering the similarity between full-precision tensor and quantized tensor. 
However, these methods all focus on applying mixed bit-widths solely to linear operations and do not consider non-linear operators. 
Our work focuses on selecting and applying mixed methods optimally for quantizing non-linear operations, unlike existing mixed quantization studies that only consider bit-width.
The proposed method is independent of the existing methods of applying mixed bit-widths and consequently has the potential to generate synergy when used simultaneously.

\begin{table}[tb]
  \caption{Previous works for non-linear quantization in Vision Transformers.}
  \label{table:non-linear approximation functions}
  \centering
  \begin{tabular}{lccccccc}
    \toprule
    Model & Softmax & {  LayerNorm  } & GELU & {  Code  }  & QAT/PTQ & {  Approximation  } \\
    \midrule
    I-ViT~\cite{Li_2023_ICCV}           & \checkmark & \checkmark & \checkmark & \checkmark & QAT & Bit-shifting \\
    I-BERT~\cite{pmlr-v139-kim21d}      & \checkmark & \checkmark & \checkmark & \checkmark & QAT & Polynomial \\
    PackQViT~\cite{dong2024packqvit}    & \checkmark & \checkmark & \checkmark &            & QAT & Polynomial \\
    FQ-ViT~\cite{Lin2021FQViTPQ}        & \checkmark &            & \checkmark & \checkmark & PTQ & Logarithm \\
    EdgeKernel~\cite{zhang2023practical}& \checkmark & \checkmark & \checkmark &            & PTQ & Logarithm \\
    SOLE~\cite{wang2023sole}            & \checkmark & \checkmark & \checkmark &            & PTQ & Logarithm \\
  \bottomrule
  \end{tabular}
\end{table}

\section{Method}
\label{sec:method}
In this section, we describe the background of the quantization, and the mixed non-linear quantization. Additionally, we explain how the layer-wise non-linear operation analysis is conducted for mixed non-linear quantization, demonstrating that the quantization error varies depending on the quantization method applied to each non-linear operation.

\subsection{Background}
\label{sec:preliminary}
In this section, we explain basic concept of the quantization used in this paper. 

\subsubsection{Basic Quantization Method and Quantization Aware Training.}
The linear operations of the vision transformer exhibit homogeneity, thus they can be quantized through linear operation quantization~\cite{jacob2018quantization, dong2020hawq} for the embedding layer (Conv), MatMul, and Dense layers. For ease of implementation, input and weight are quantized using symmetric uniform manner as described in Eq.~\eqref{eq:unifrom_q}. In Eq.~\eqref{eq:unifrom_q}, $R$ represents the full-precision value, and $\alpha$ denotes the clipping range, which is determined by the larger of the absolute values of $r_{min}$ and $r_{max}$. $n$ represents the bit-precision.
\begin{equation}
\label{eq:unifrom_q}
I = \left\lfloor \frac{\text{clip}(R, -\alpha, \alpha)}{S} \right\rceil, \quad \text{where} ; S = \frac{2\alpha}{2^n - 1}, \alpha = \text{max}(|r_{max}|, |r_{min}|)
\end{equation}

To address the issue where gradients cannot back-propagate in Eq.~\eqref{eq:unifrom_q}, the Straight-Through Estimator (STE)~\cite{bengio2013estimating} is used to approximate the gradient during quantization-aware training. The gradient of the rounding operation is approximated as 1 within the quantization limit. 
In back-propagation using STE, the gradient of the loss $\mathcal{L}$ with respect to the real-valued data $R$ is given by Eq.~\eqref{eq:ste}

\begin{equation}
    \frac{\partial \mathcal{L}}{\partial R} = \frac{\partial \mathcal{L}}{\partial I} \cdot \mathbf{1}_{-\alpha \leq \frac{R}{S} \leq \alpha},
\label{eq:ste}
\end{equation}

where $\mathbf{1}$ functions as an indicator, yielding 1 within the boundaries of quantization and 0 outside these boundaries. 

\subsubsection{Non-linear Quantization Methods.}
A vision transformer contains three non-linear operations: layer normalization (LayerNorm), Softmax, and GELU.
Studies have proposed approximating each of these non-linear operations with specific integer-only functions, as shown in Table~\ref{table:non-linear approximation functions}.
In this study, we utilize three techniques that are open-source and easily accessible: I-BERT~\cite{pmlr-v139-kim21d}, FQ-ViT~\cite{Lin2021FQViTPQ}, and I-ViT~\cite{Li_2023_ICCV}, applying a mixed approach to each non-linear operation.
The quantization methods for non-linear operations are detailed as follows.

LayerNorm includes division, square, and square root as shown in Eq.~\eqref{equation:LayerNorm}. 
The computation of the mean and standard deviation is dynamically calculated based on the input values, which presents challenges.
\begin{equation}
\label{equation:LayerNorm}
\text{LayerNorm}(X) = \frac{X - \mu_X}{\sqrt{\sigma_X^2 + \epsilon}} \cdot \gamma + \beta
\end{equation}
Non-linear quantizations of I-BERT and I-ViT compute exact value of $\sqrt{n}$ using an iterative algorithm based on Newton's Method~\cite{crandall2005prime}.
In this process, Newton's Method requires only integer arithmetic operations.
Quantization of I-ViT differs from I-BERT in that it represents the division by 2 of the sum of the i-th iteration value and the rounded value with a bit-shift, and the number of iterations is determined empirically.
In FQ-ViT, LayerNorm is approximated to integers with the output scaling factor obtained from PTQ along with the log2 and the sign function.

In Softmax, as shown in Eq.~\eqref{equation:Softmax}, the exponential function introduces non-linearity, making it crucial to approximate. 
I-ViT applies the base change formula to transform the base of the exponential function from $e$ to $2$, enabling shift operations. 
As shown in Eq.~\eqref{equation:base changing formula}, where the $\log_2 e$ appearing in the exponent is approximated as $(1.0111)_b$, allowing the binary value to be represented by shift operations and addition. 
I-BERT approximates the exponential function with a second-order polynomial. To determine the coefficients of the polynomial, i-exp~\cite{pmlr-v139-kim21d} is proposed, which minimizes the $L^2$ distance in the exponential function. FQ-ViT proposes Log-Int-Softmax, which adds log2 quantization to the i-exp of I-BERT.

\begin{equation}
\label{equation:Softmax}
\text{Softmax}(x_i) = \frac{e^{x_i}}{\sum_{j}^{d} e^{x_j}}
\end{equation}
\begin{equation}
\label{equation:base changing formula}
e^{x} = 2^{x \cdot \log_2 e} \approx 2^{x \cdot (1 + (1 \gg 1) - (1 \gg 4))}
\end{equation}

The GELU function poses quantization issues with the erf function as in Eq.~\eqref{equation:GELU}. I-BERT proposed i-GELU by approximating the erf function with a second order polynomial function. I-ViT employs an approximation based on the sigmoid proposed by GELUs~\cite{hendrycks2016gaussian}, as in Eq.~\eqref{equation:h-GELU}.
\begin{equation}
\label{equation:GELU}
\text{GELU}(x) = x \cdot \frac{1}{2} \left[1 + \text{erf}\left(\frac{x}{\sqrt{2}}\right)\right]
\end{equation}

\begin{equation}
\label{equation:h-GELU}
\text{GELU}(x) \approx {x \cdot \sigma(1.702x)} 
\end{equation}
The value 1.702 in Eq.~\eqref{equation:h-GELU} is approximated as the binary $(1.1011)_b$ and represented with the shift operation as shown in Eq.~\eqref{equation:1.702}. The sigmoid function uses ShiftExp, which was employed for quantizing LayerNorm, to quantize the exponential function, resulting in the proposed ShiftGELU.
\begin{equation}
\label{equation:1.702}
{x \sigma(1.702x)} = x \sigma(x + (x \gg 1) + (x \gg 3) + (x \gg 4))
\end{equation}

\subsection{Mixed Non-linear Quantization}%
\label{sec:Mixed Non-linear Quantization}
In this paper, we propose a method to select the optimal non-linear quantization method for each non-linear layer. This is implemented exploiting the distribution of output activation and input tensor. We calculate the quantization error for each layer to choose the quantization method for non-linear operations. We measure layer-wise quantization sensitivity of the three non-linear quantization methods, I-BERT, FQ-ViT, and I-ViT, and select the quantization method that induces the minimum error. For GELU, since FQ-ViT does not provide a quantization method, there are only two available options. For Softmax and LayerNorm, there are three available methods (I-BERT, FQ-ViT, and I-ViT), and the final search space of the non-linear quantization becomes Eq.~\eqref{eq:searchSpace}.
\begin{equation}
\label{eq:searchSpace}
3^{\text{(\# of Softmax)}} \times 3^{\text{(\# of LayerNorm)}} \times 2^{\text{(\# of GELU})}
\end{equation}
In the case of the ViT, which contains 12 Softmax, 12 GELU, and 25 LayerNorm, the size of the search space becomes approximately $9.47 \times 10^{18}$. Calculating all dependencies among non-linear operations is intractable. Therefore, like previous studies, this study also assumes that all layers are independent~\cite{dong2019hawq,dong2020hawq,yao2021hawq}.
In this case, only $3\times12+3\times12+2\times25 = 122$ computations are required.

The proposed mixed non-linear quantization progresses through the following three stages:
i) Quantized model setting, ii) Calculate layer-wise quantization sensitivity, and iii) Selecton of the optimal non-linear quantization method.

\subsubsection{Quantized Model Setting.}
To calculate layer-wise quantization sensitivity for mixed non-linear quantization, models quantized with three non-linear quantization methods~\cite{Li_2023_ICCV, Lin2021FQViTPQ, pmlr-v139-kim21d} are prepared.
The linear quantization part follows the Dyadic Quantization of HAWQv3~\cite{yao2021hawq}, and the base code from the official I-ViT~\cite{Li_2023_ICCV} was used.
Only the non-linear quantization part of the base code was implemented differently using the three methods to configure the quantized models.

\subsubsection{Calculate Layer-wise Quantization Sensitivity.}
We perform inference on both full-precision vision transformer and quantized vision transformers, and generate a list of quantization sensitivity of each layer. The Signal-to-Quantization-Noise Ratio (SQNR) is used to measure the quantization sensitivity.
SQNR represents the ratio between the original data and the quantization noise.
We measure the quantization error induced from rounding during the quantization process by comparing the original tensor with the tensor after quantization and dequantization.
A higher SQNR value indicates that the quantized non-linear layer produces a tensor similar to that of the full-precision non-linear layer, implying excellent integer approximation accuracy of the layer.
In Eq.~\eqref{equation:ASQNR}, $x$ represents the original tensor, and $q$ represents the tensor after quantization and dequantization.
Therefore, quantization error is calculated through $x-q$.
For the final SQNR calculation, the original tensor is divided by the quantization error and the ratio is converted into a $log scale$.
The calculation of SQNR is ultimately done for a batch size $N$, and thus it is computed as the arithmetic mean accumulated over the batch. This average SQNR(ASQNR) is defined as Eq.~\eqref{equation:ASQNR}
This process is carried out for all layers in the ViT model.
Finally, layer-wise quantization sensitivity for the model quantized with the three considered quantization methods is generated.

\begin{equation}
\label{equation:ASQNR}
\text{ASQNR}  = 20 \log \left( \frac{1}{N} \sum_{i=1}^{N} \frac{\mathbb{E}[(x_i)^2]}{\mathbb{E}[(x_i - q_i)^2]} \right)
\end{equation}

\subsubsection{Selection of the optimal non-linear quantization method.}
In the previous stage, three list of layer-wise quantization sensitivities were generated, allowing for the selection of the optimal non-linear quantization for each non-linear layer. 
This paper does not solely choose the quantization method based on the output tensor's ASQNR, but utilizes the newly devised \textit{SQNR diff} metric. 
\textit{SQNR diff} is the difference between the quantization sensitivities of the current non-linear operation's input and output tensors, thus considering the quantization sensitivity of both input and output. 
As a criterion for selecting the non-linear quantization method, the method with the lowest \textit{SQNR diff} for each non-linear layer is chosen for the quantization of that non-linear layer. 
Experimental comparisons between the \textit{SQNR diff}-based selection proposed in this paper and the method of choosing quantization based on high output quantization sensitivity (\textit{SQNR output}) showed that \textit{SQNR diff} is more effective in improving accuracy. 
The results of accuracy comparison between the two assignment methods are summarized in Table~\ref{table:Compared with two decision rules}.

\begin{equation}
\label{equation:SQNR diff}
\begin{aligned}
& \text{SQNR diff} = \beta - \gamma \\ 
& \text{where} ; \beta  = \text{ASQNR}(X_{out}, Q_{out}), \gamma = \text{ASQNR}(X_{in}, Q_{in}) \\
\end{aligned}
\end{equation}

\textit{SQNR diff} is calculated as shown in Eq.~\eqref{equation:SQNR diff}.
In Eq.~\eqref{equation:SQNR diff}, $X_{out}$ and $X_{in}$ refer to the output and input tensors of the non-linear layer in the full-precision model, respectively, and $Q_{out}$ and $Q_{in}$ denote the output and input tensors that represent the tensor after quantization and dequantization. 
The \textit{SQNR diff} can be calculated by subtracting the input quantization sensitivity $\gamma$ from the output quantization sensitivity $\beta$ generated after passing through the quantized non-linear layer.

\begin{table*}
\centering
\caption{Accuracy comparison of two decision rules for the 8-bit mixed non-linear quantization in vision transformers. The \textit{SQNR diff} rule achieved higher accuracy than SQNR output decision rule. All models were trained in 5 epochs and the same batch size for each model.}
\label{table:Compared with two decision rules}
\resizebox{\textwidth}{!}{
\begin{tabular}{lcccccccc}
\toprule
\multicolumn{1}{l}{\textbf{Decision rule}} &
\multicolumn{1}{c}{\textbf{ ViT-S }} &
\multicolumn{1}{c}{\textbf{ ViT-B }} &
\multicolumn{1}{c}{\textbf{ DeiT-T }} &
\multicolumn{1}{c}{\textbf{ DeiT-S }} &
\multicolumn{1}{c}{\textbf{ DeiT-B }} &
\multicolumn{1}{c}{\textbf{ Swin-T }} & 
\multicolumn{1}{c}{\textbf{ Swin-S }} \\
\midrule
\midrule
SQNR output     
& 63.01 & 34.27 & 70.45 & 79.74 & 81.19 & 5.81 & 79.49 \\
SQNR diff  
& 79.28 & 83.25 & 71.88 & 79.78 & 81.65 & 79.32 & 80.67 \\
\bottomrule
\end{tabular} }
\end{table*}

\begin{figure}[t]
    \centering
    \includegraphics[height=11cm, width=\linewidth]{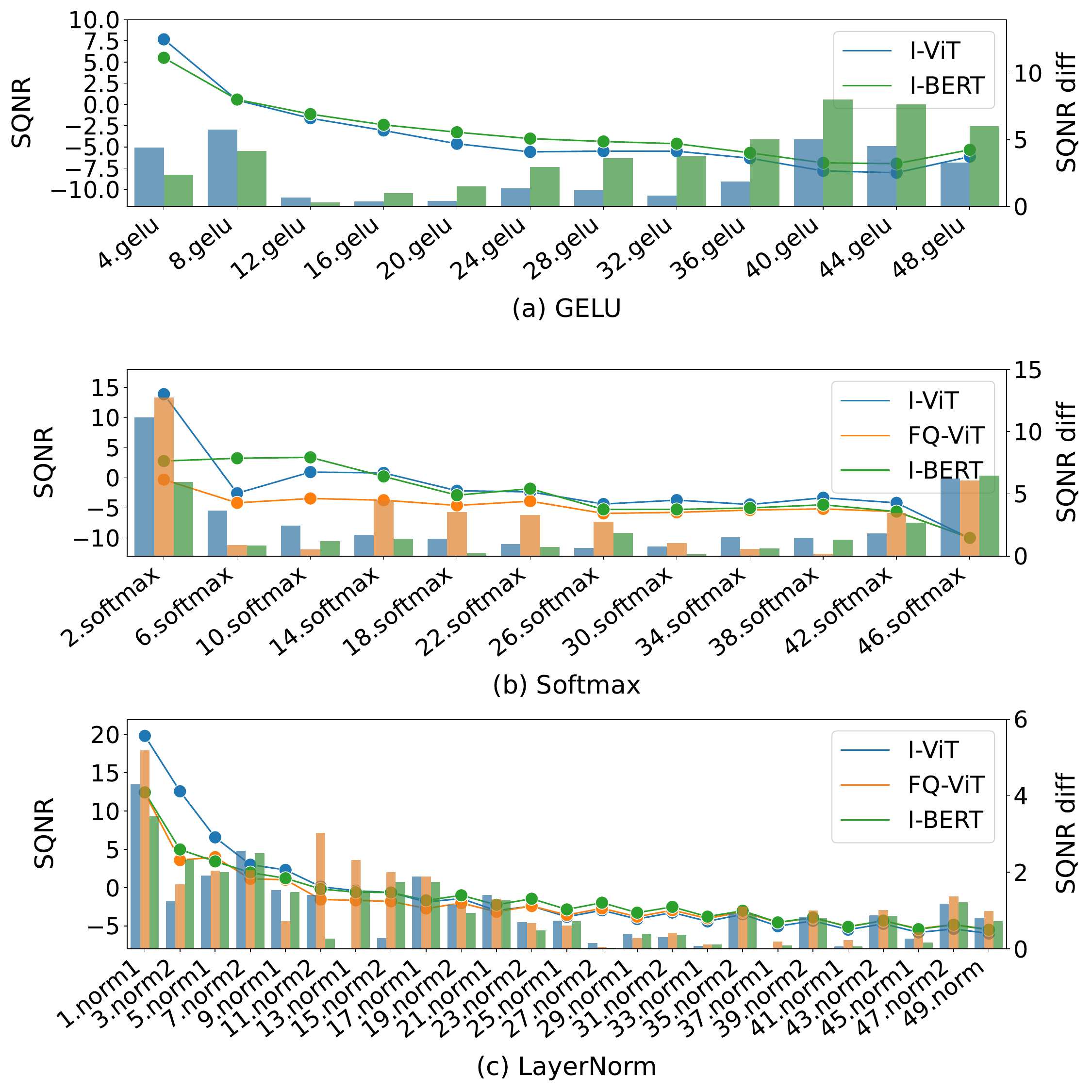}
    \caption{Layer-wise quantization sensitivity of non-linear operations for quantized DeiT-T with the existing non-linear quantization methods. Bar plots denotes \textit{SQNR diff} of each non-linear layer, while the line plots denote \textit{SQNR} output. Bar plots indicate no single quantization method uniformly minimizes \textit{SQNR diff} across all non-linear layers.}
    \label{fig:table_sqnr_layers_DeiT}
\end{figure}

\begin{figure}[t]
  \centering
  \includegraphics[height=7.5cm, width=\linewidth]{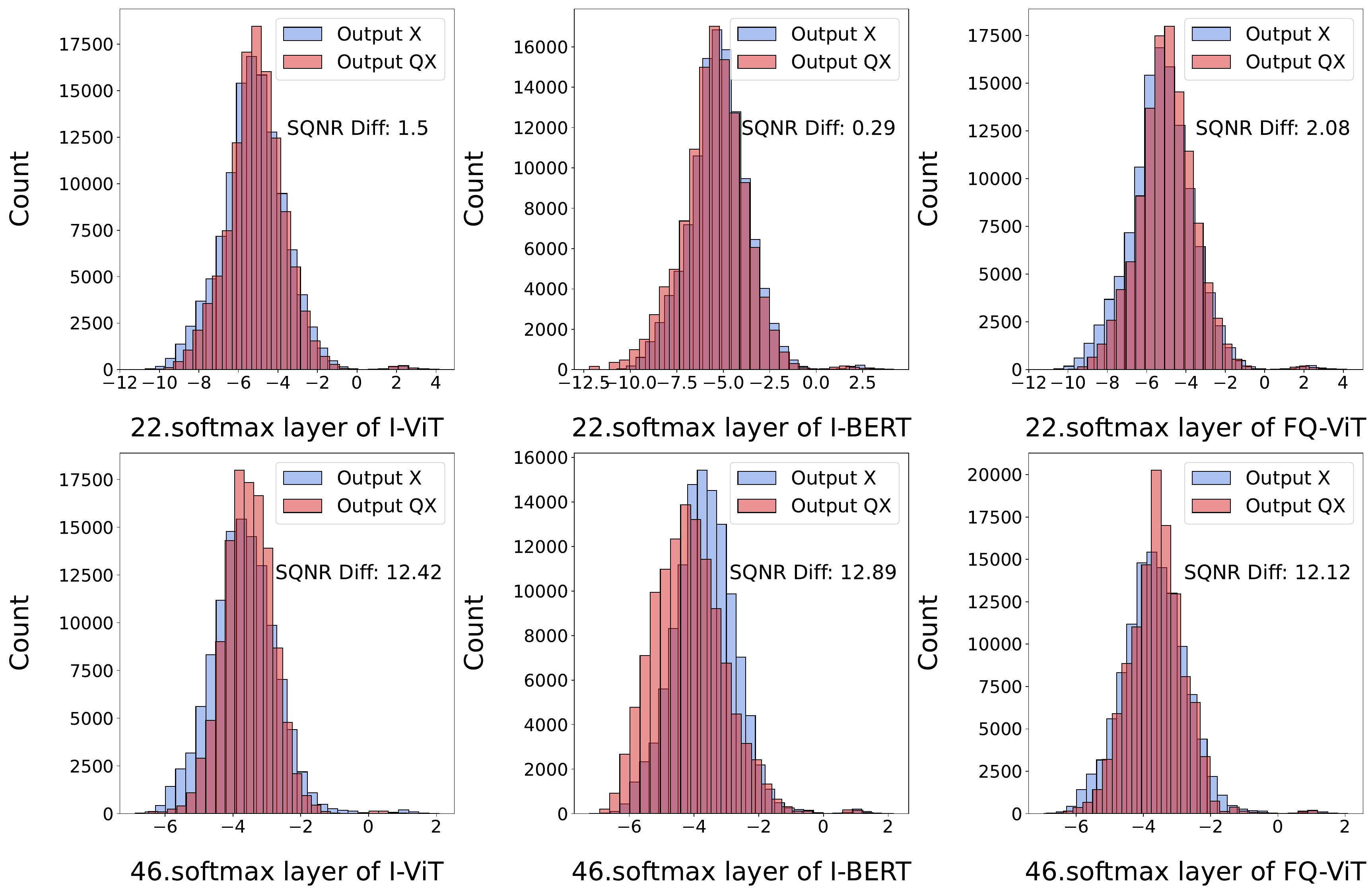}
  \caption{Histogram comparison of output tensor for specific non-linear layers in full-precision and quantized DeiT-T models. There is no effective non-linear quantization method which approximate full-precision output tensor X for all non-linear layers. Each subplot shows the distribution of full-precision (\textit{Output X}) and quantized (\textit{Output QX} after quantization and dequantization) tensor, along with their \textit{SQNR diff}. The layer names are in accordance with Figure~\ref{fig:table_sqnr_layers_DeiT}.}
  \label{fig:deit-t-histogram}
\end{figure}

\section{Experiments}
\label{sec:experiment}
In this section, we describe experiment environment, results of layer-wise non-linear operation analysis and performance evaluation results.
SQNR analysis related to the position of non-linear operations and the resulting non-linear quantization mappings is performed.
Effectiveness of the proposed mixed non-linear quantization is shown by implementing onto the well known vision transformers. Various non-linear quantizations are compared in 8-bit and 6-bit environment. Efficiency from a training time perspective is discussed.

\subsection{Experiment Setup}%
\label{sec:implementation details}
ViT~\cite{dosovitskiy2020image}, DeiT~\cite{touvron2021training}, and Swin~\cite{liu2021swin} are selected to implement the mixed non-linear quantization with the ImageNet dataset~\cite{krizhevsky2012imagenet}.
All these models utilized pre-trained weights downloaded from the timm\footnote{\url{https://github.com/huggingface/pytorch-image-models}} library, and Quantization Aware Training (QAT) was conducted for quantization. 
Uniform symmetric quantization was applied to quantize all weights and activations of the full-precision pre-trained vision transformer models, and basic QAT based on STE~\cite{bengio2013estimating} was performed. 
The optimizer used was AdamW. 
Four A6000 and four A100 GPUs were utilized for sensitivity analysis and QAT.
The GPU training time for QAT, based on the A100-80G, took an average of 7 days for the Tiny, Small, and Base models.

\subsection{Layer-Wise Non-linear Operation Analysis}%
\label{sec:layer-wise non-linear operation analysis results}
In this section, we experimentally demonstrate that both SQNR and \textit{SQNR diff} vary depending on the layer position of the non-linear operation in the quantization process. Additionally, we present the mapping results of the proposed \textit{SQNR diff}-based mixed non-linear quantization for each model.

Figure~\ref{fig:table_sqnr_layers_DeiT} shows the SQNR and \textit{SQNR diff} values for the non-linear operations of the DeiT-T model, quantized by different non-linear quantization methods(I-ViT, FQ-ViT and I-BERT). The quantization sensitivity of the output tensor for GELU, Softmax, and LayerNorm is represented in a line plot, while the measurements taken using the proposed \textit{SQNR diff} metric are displayed in a bar plot. Despite applying the same non-linear quantization to the same operations in the non-linear layers, different quantization sensitivities can be observed in Figure~\ref{fig:table_sqnr_layers_DeiT}. For example, in the line plot for Softmax, the \textit{2.softmax} layer showed superior performance with I-ViT, while the \textit{6.softmax}, \textit{10.softmax} and \textit{22.softmax} layers exhibited higher quantization sensitivity values with I-BERT, and the \textit{46.softmax} layer did so with FQ-ViT. The individual \textit{SQNR diff} for each non-linear layer, the bar plot in Figure~\ref{fig:table_sqnr_layers_DeiT} confirmed that the \textit{SQNR diff} values varied among all layers. Therefore, no single quantization method demonstrated superior performance across all non-linear layers.

In the Figure~\ref{fig:table_sqnr_layers_DeiT}, the effectiveness of the quantization performance varied across different positions of the Softmax layers. The original output tensor and the quantized output tensor are analyzed in a histogram as shown in Figure~\ref{fig:deit-t-histogram}. In the histogram, the blue area labeled \textit{Output X} represents the output tensor of the layer from the full-precision model, and the red area labeled \textit{Output QX} represents the output tensor that has passed through the layer with non-linear quantization. 
For the \textit{22.Softmax}, the I-BERT has the smallest \textit{SQNR diff=0.29} and for the \textit{46.Softmax}, the FQ-ViT has the smallest \textit{SQNR Diff=12.12} so that they are the non-linear quantization methods that achieve the highest preservation of the original tensors.
Since the optimal non-linear method varies depending on the activation distributions, applying a mixed approach to non-linear quantization can be effective in reducing quantization errors.

From the quantization sensitivity analysis for non-linear operations, non-linear quantization techniques can be mixed and applied to each non-linear layer.
As shown in Table~\ref{table:Compared with two decision rules}, the \textit{SQNR diff} decision rule demonstrated higher accuracy than the output decision rule in all models. Therefore, the proposed mixed non-linear quantization method, based on \textit{SQNR diff} values, was applied to the ViT, DeiT, and Swin models, and the final mixed mapping results are presented in Table~\ref{table:Softmax_gelu_norm_counts}. These results indicate that no single method consistently outperforms others across the individual non-linear operations of a single model.

\begin{table*}[t]
\centering
\caption{This table shows which and how much of non-linear quantization was selected for each non-linear layer with the mixed non-linear quantization.}
\label{table:Softmax_gelu_norm_counts}
\resizebox{\textwidth}{!}{%
\begin{tabular}{lccc|cc|ccc}
\toprule
\multirow{2}{*}{\textbf{Model}} & \multicolumn{3}{c}{\textbf{Softmax}} & \multicolumn{2}{c}{\textbf{GELU}} & \multicolumn{3}{c}{\textbf{LayerNorm}} \\



& \textbf{I-BERT}
& \textbf{FQ-ViT}
& \textbf{I-ViT}

& \textbf{I-BERT}
& \textbf{I-ViT}

& \textbf{I-BERT}
& \textbf{FQ-ViT}
& \textbf{I-ViT} \\
\midrule
\midrule
ViT-S  & 1 & 7 & 4 & 12 & 0 & 2 & 0 & 23 \\
\midrule
ViT-B  & 1 & 8 & 3 & 12 & 0 & 0 & 1 & 24 \\
\midrule
DeiT-T & 6 & 4 & 2 & 3 & 9 & 13 & 4 & 8 \\
\midrule
DeiT-S & 8 & 1 & 3 & 11 & 1 & 23 & 1 & 1 \\
\midrule
DeiT-B & 11 & 0 & 1 & 11 & 1 & 2 & 0 & 23 \\
\midrule
Swin-T & 5 & 0 & 7 & 11 & 1 & 18 & 4 & 3 \\
\midrule
Swin-S & 6 & 15 & 3 & 18 & 6 & 11 & 27 & 11 \\
\bottomrule
\end{tabular}%
}
\end{table*}


\begin{table*}[t]
\centering
\caption{Top-1 accuracy (\%) of quantized 8-bit and 6-bit vision transformer models. ${}^{\ast}$ denotes the results that are re-produced with the official code. }
\label{table:Comprehensive Results Table 1}
\resizebox{\textwidth}{!}{
\begin{tabular}{lccccccccc}
\toprule
\textbf{Method} & 
\textbf{Bits} & 
\textbf{ViT-S} & 
\textbf{ViT-B} & 
\textbf{DeiT-T} & 
\textbf{DeiT-S} & 
\textbf{DeiT-B} & 
\textbf{Swin-T} & 
\textbf{Swin-S} \\
\midrule
\midrule
Full-Precision               
& 32-bit & 81.39 & 84.53 & 72.21 & 79.85 & 81.85 & 81.35 & 83.2 \\
\midrule
FQ-ViT~\cite{Lin2021FQViTPQ}    & 8-bit & --   & 83.31 & 71.61 & 79.17 & 81.20 & 80.51 & 82.71 \\
I-BERT~\cite{pmlr-v139-kim21d}  & 8-bit & 80.47 & 83.70 & 71.33 & 79.11 & 80.79 & 80.15 & 81.86 \\
I-ViT~\cite{Li_2023_ICCV}       & 8-bit & 81.27 & 84.76 & 72.24 & 80.12 & 81.74 & 81.50 & 83.01 \\
\textbf{Ours}                   & 8-bit & \textbf{81.35} & \textbf{84.88} & \textbf{72.55} & \textbf{79.93} & \textbf{81.88} & \textbf{80.95} & \textbf{82.42} \\
\midrule

FQ-ViT~\cite{Lin2021FQViTPQ}                    & 6-bit & 4.26 & 0.10 & 58.66 & 45.51 & 64.63 & 66.50 & 52.09 \\
I-ViT${}^\ast$~\cite{Li_2023_ICCV}              & 6-bit & 70.24  & 76.89 & 63.80 & 74.48 & 76.0 & 71.89 & 81.04 \\
\textbf{Ours}                                   & 6-bit & \textbf{73.64} & \textbf{82.00} & \textbf{67.81} & \textbf{77.19} & \textbf{79.91} & \textbf{80.54} & \textbf{79.10} \\
\bottomrule
\end{tabular}
}
\end{table*}

\subsection{Accuracy Evaluation}
\label{sec:accuracy evaluation}
Experiments were conducted comparing the proposed mixed non-linear quantization method with previous studies I-BERT, FQ-ViT, and I-ViT in terms of 8-bit and 6-bit quantization accuracy on ViT, DeiT, and Swin models. The results are presented in Table~\ref{table:Comprehensive Results Table 1}, showing an overall improvement in accuracy compared to existing quantized models, with ViT-S, ViT-B, DeiT-T, DeiT-S, and DeiT-B models even surpassing their full-precision counterparts. Specifically, the DeiT-T model achieved an accuracy of 72.55\% with our 8-bit mixed non-linear quantization, which is 0.34\%p higher than the full-precision model. It also exceeded the accuracy of I-ViT by 0.31\%p, I-BERT by 1.22\%p, and FQ-ViT by 0.94\%p. For the 6-bit quantization comparison, results were reproduced and compared using official codes after sufficient retraining. The proposed method improved accuracy by an average of 35.5\%p and 19.6\%p over FQ-ViT and I-ViT, respectively.

\begin{figure}[tb]
  \centering
  \includegraphics[height=6cm, width=0.9\linewidth]{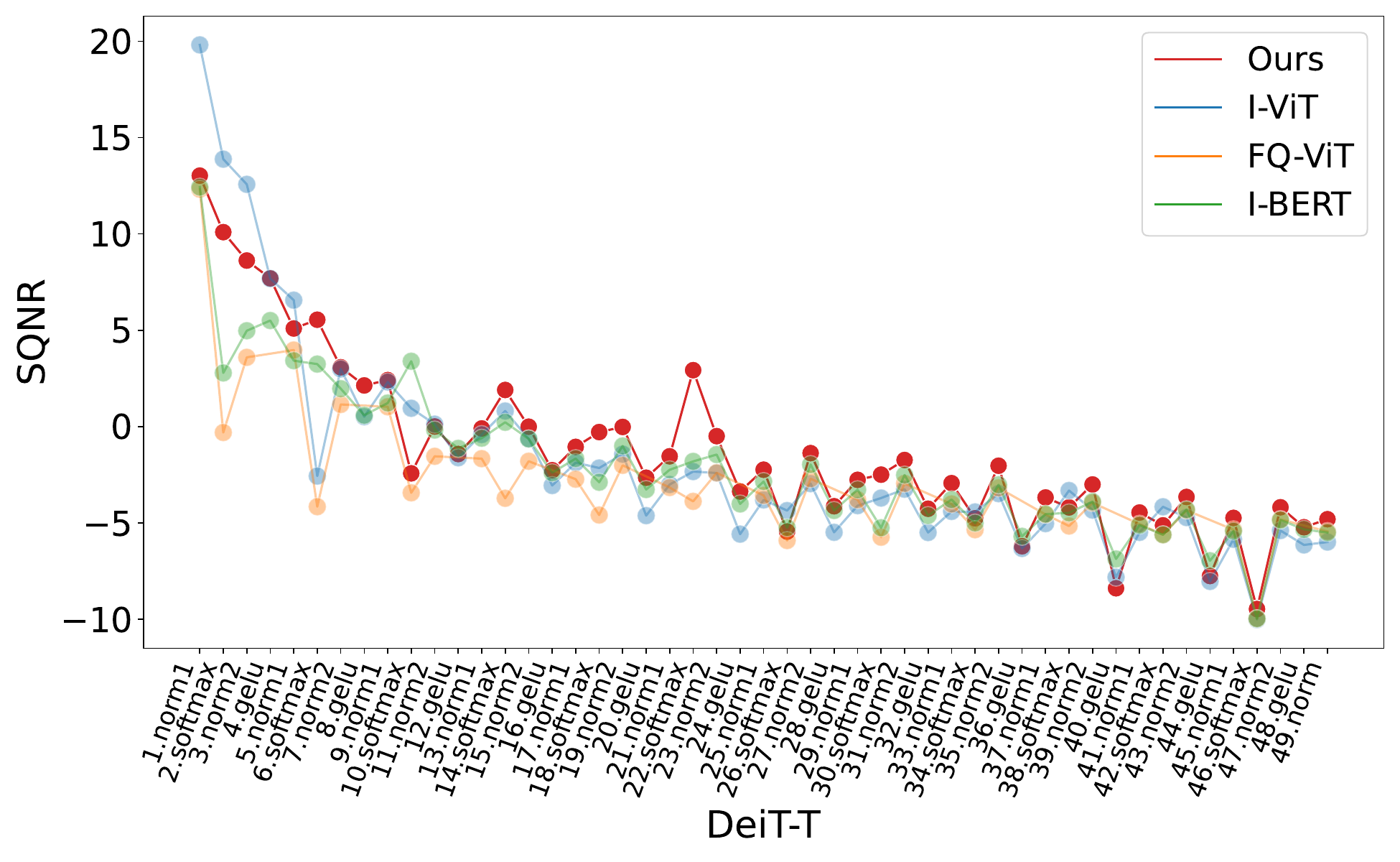}
  \caption{Layer-wise quantization sensitivity of non-linear operations for quantized DeiT-T model, our method (red) improves overall quantization sensitivity for each non-linear layers.}
  \label{fig:sqnr_layers_o_DeiT_T}
\end{figure}

The SQNR of each quantized non-linear layer after QAT is shown in the Figure~\ref{fig:sqnr_layers_o_DeiT_T}. Experimental results indicate that, unlike previous methods where the SQNR values of the same non-linear operations decrease with increasing layer depth, leading to increased quantization errors, our proposed method maintains the highest SQNR in overall. This is due to our \textit{SQNR diff}-based selection approach for each non-linear layer, which reduces the cumulative quantization error as layers deepen.


%

\subsection{Training Time}%
\begin{figure}[tb]
  \centering
  \begin{subfigure}{0.49\linewidth}
    \centering
    \includegraphics[width=\linewidth]{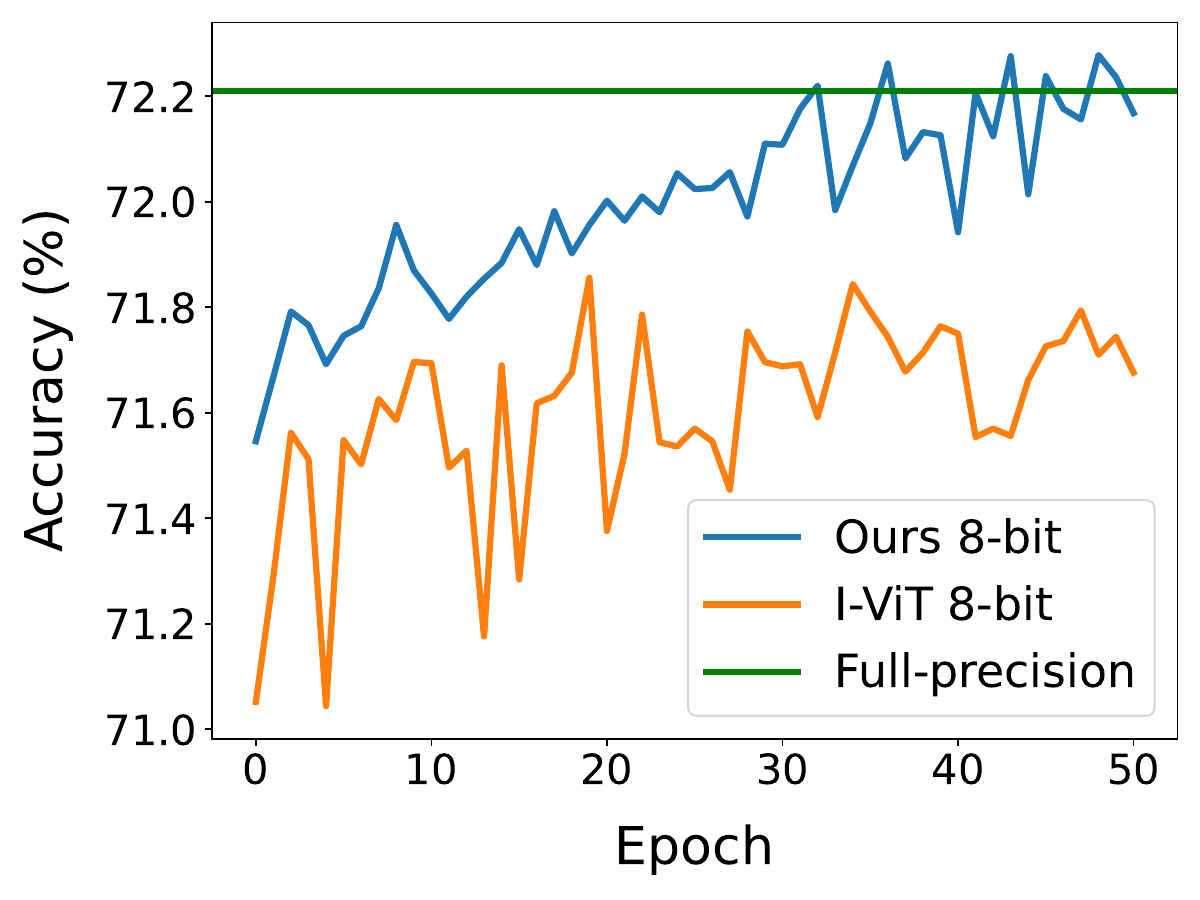}
    \caption{Training curve of 8bit DeiT-T}
    \label{fig:training_time_8bit}
  \end{subfigure}
  \begin{subfigure}{0.49\linewidth}
    \centering
    \includegraphics[width=\linewidth]{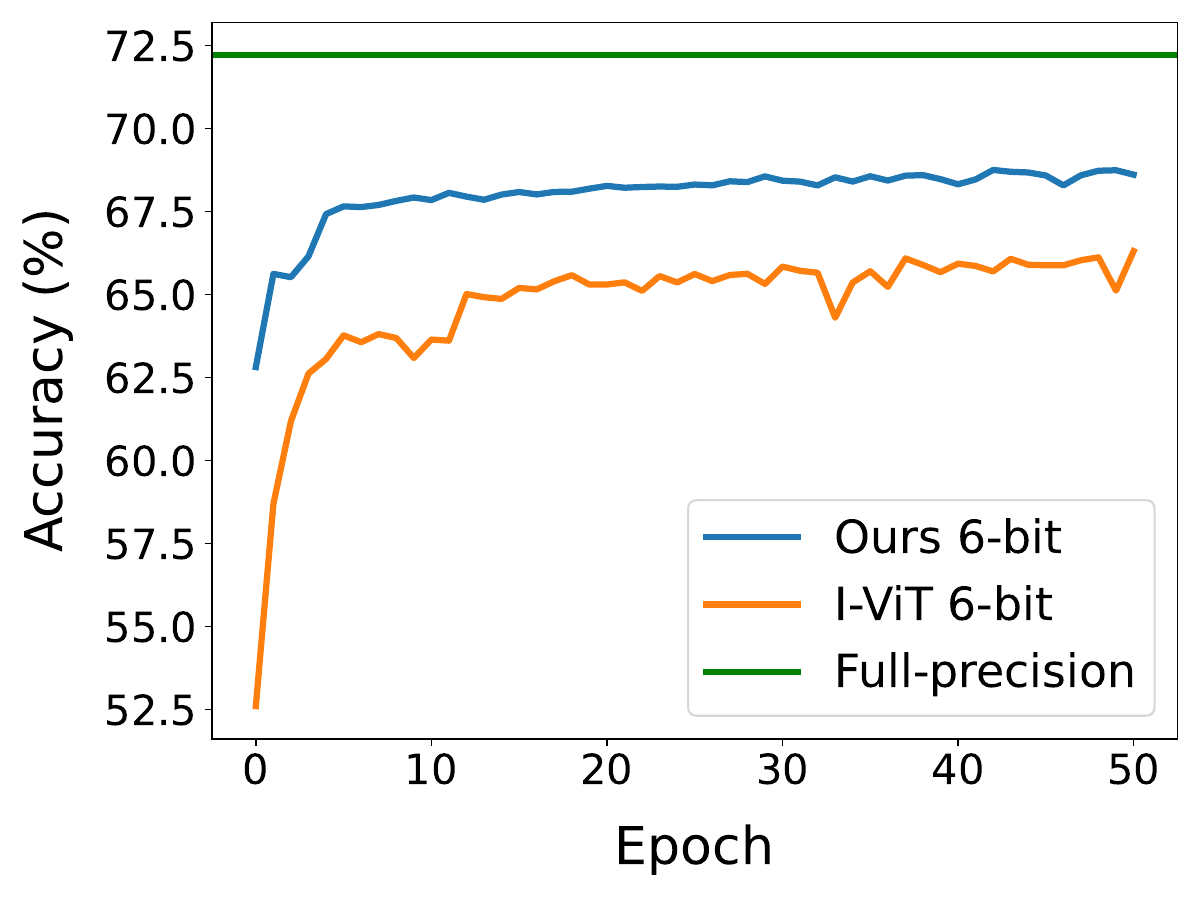}
    \caption{Training curve of 6bit DeiT-T}
    \label{fig:training_time_6bit}
  \end{subfigure}
  \caption{Training curve of DeiT-T 6-bit and 8-bit models. Our method achieves faster training convergence and higher final accuracy than I-ViT QAT.}
  \label{fig:training_time}
\end{figure}

\label{sec:Training Time}
The mixed non-linear quantization method is evaluated for its learning efficiency by measuring training time to achieve the accuracy with the QAT. 
For comparison, training time was measured against the most recent studies on I-ViT QAT with the official code. Experimental results are shown in Figure~\ref{fig:training_time}, which indicates that at the same epoch, our method achieved higher accuracy in both 8-bit and 6-bit quantization, showing not only faster convergence but also higher final accuracy.

For rapid deployment, the accuracy improvement when applying QAT for only one epoch is shown in Table~\ref{table:1-epoch training}. Mixed non-linear quantization achieved higher Top-1 accuracy across all models compared to I-ViT and I-BERT. As for performance drop from full training to only one epoch of training, I-ViT showed an average accuracy decrease of -22.27\%p, whereas the proposed method only reduced accuracy by an average of -1.33\%p compared to the 8-bit quantization results of the Table~\ref{table:Comprehensive Results Table 1}. Thus, the proposed method demonstrates that it can quickly correct accuracy even when QAT is applied for just one epoch, making it suitable for rapid deployment.

\begin{table*}[t]
\centering
\caption{1-epoch training results for 8-bit quantization. Our method shows faster training speed then I-ViT and I-BERT. ${}^{\ast}$ indicates the results that are re-produced by paper I-ViT~\cite{Li_2023_ICCV}.}
\label{table:1-epoch training}
\resizebox{\textwidth}{!}{
\begin{tabular}{lcccccccc}
\toprule
\multicolumn{1}{l}{\textbf{Method}} &
\multicolumn{1}{c}{\textbf{Bit-prec.}} &
\multicolumn{1}{c}{\textbf{ ViT-S }} &
\multicolumn{1}{c}{\textbf{ ViT-B }} &
\multicolumn{1}{c}{\textbf{ Deit-T }} &
\multicolumn{1}{c}{\textbf{ Deit-S }} &
\multicolumn{1}{c}{\textbf{ Deit-B }} &
\multicolumn{1}{c}{\textbf{ Swin-T }} & 
\multicolumn{1}{c}{\textbf{ Swin-S }} \\
\midrule
\midrule
I-ViT~\cite{Li_2023_ICCV}                   & 8bit & 57.20 & 30.53 & 71.05 & 78.27 & 28.93 & 77.65 & 65.13 \\
I-BERT${}^{\ast}$~\cite{pmlr-v139-kim21d}   & 8bit & 76.63 & 81.62 & 70.81 & 79.17 & 81.35 & 80.54 & 80.63 \\
Ours                                        & 8bit & 77.68 & 82.55 & 71.54 & 79.24 & 81.67 & 80.52 & 81.43 \\
\bottomrule
\end{tabular} }
\end{table*}

\section{Conclusion}
In this paper, we proposed a mixed non-linear quantization method that considers the sensitivity of each non-linear layer. This method strategically assigns the most error-minimizing non-linear quantization method from the known non-linear quntization operations.
We also devised a novel SQNR difference metric, \textit{SQNR diff}, to accurately assess layer-wise quantization effects.
Experimental results demonstrated that our method significantly improved accuracy, outperforming existing methods such as I-BERT, FQ-ViT, and I-ViT in both 8-bit and 6-bit environments. Even under the constraint of a single epoch training time for rapid deployment, our approach still yielded higher accuracy improvements over existing methods.

\section*{Acknowledgment}
This work was supported by Institute of Information \& communications
Technology Planning \& Evaluation (IITP) grant funded by the Korea
government (MSIT) (No.RS-2023-00277060, Development of open edge AI SoC hardware and software platform).
This work was supported by Korea Research Institute for defense Technology planning and advancement (KRIT) grant funded by the Korea government (DAPA(Defense Acquisition Program Administration)) (No.KRIT-CT-22-040, Heterogeneous Satellite constellation based ISR Research Center, 2024).

\bibliographystyle{splncs04}
\bibliography{main}

\end{document}